\def\year{2022}\relax
\documentclass[letterpaper]{article} 
\usepackage{aaai22}  
\usepackage{times}  
\usepackage{helvet}  
\usepackage{courier}  
\usepackage[hyphens]{url}  
\usepackage{graphicx} 
\usepackage{comment}
\usepackage{natbib}  
\usepackage{caption} 
\DeclareCaptionStyle{ruled}{labelfont=normalfont,labelsep=colon,strut=off} 
\usepackage{algorithm}
\usepackage{algorithmic}

\usepackage{newfloat}
\usepackage{listings}
\floatstyle{ruled}
\newfloat{listing}{tb}{lst}{}
\floatname{listing}{Listing}
\begin{document}
\title{Is AI Art Another Industrial Revolution in the Making?}
\author{Alexis Newton, Kaustubh  Dhole\\
}
\affiliations{
    Department of Computer Science\\
    Emory University\\
    \{annewto, kdhole\}@emory.edu
%
}
\maketitle
\begin{abstract}
A major shift from skilled to unskilled workers was one of the many changes caused by the Industrial Revolution, when the switch to machines contributed to decline in the social and economic status of artisans, whose skills were dismembered into discrete actions by factory-line workers. We consider what may be an analogous computing technology: the recent introduction of AI-generated art software. AI art generators such as Dall-E and Midjourney can create fully rendered images based solely on a user’s prompt, just at the click of a button. Some artists fear if the cheaper price and conveyor-belt speed that comes with AI-produced images is seen as an improvement to the current system, it may permanently change the way society values/views art and artists.
In this article, we consider the implications that AI art generation introduces through a post-industrial revolution historical lens. We then reflect on the analogous issues that appear to arise as a result of the AI art revolution, and we conclude that the problems raised mirror those of industrialization, giving a vital glimpse into what may lie ahead.

\end{abstract}

\section{Introduction}
The industrial revolution caused a major shift from skilled to unskilled labor when machines contributed to a massive layoff of artisans in favor of factory-line workers. William Pelz describes how prior to the industrial revolution people had been working in much the same way for thousands of years, producing goods through human labor with some assistance via animals or water power. However, he says, “All of this changed with the rise of the machine: tools would no longer serve people, but rather people would serve machines” \cite{pelz}. Pelz argues that due to this massive change in goods production, the people became an “appendage” to the machine, as humans were suddenly at the mercy of machines using them to make goods. As factory life took hold, compensation was no longer based on paying for skill, but rather on paying for time. This helped to usher in a different kind of day-to-day experience for most commoners--that of a time-work discipline. Maxine \citet{berg} notes the change to a mechanized factory sector from pre-industrial handicrafts. Many traditional workers had to transition to this type of work, or risk being left behind by a quickly transforming industry, leading to the devaluation and destruction of businesses that could not compete with the cheaper and faster production that industrialization had to offer.

With the recent introduction of AI-based art models \cite{ho2020denoising, dalle2, ruiz2022dreambooth}, we argue that a shift largely reminiscent of the post-industrial revolution is unfolding. As AI-based models become more and more common, issues artisans experienced in the mid 1800s are reemerging, which similarly question the very existence of artists today. As we are on the precipice of this revolution, it is imperative for all stakeholders, viz, policymakers, ethicists, computer scientists and artists to understand what such a shift would entail in order to manage the consequences of that shift.

In this article, our aim is to view recent developments in the art industry due to the introduction of artificial intelligence models through a post-industrial lens. In the following sections, we first discuss how technologies influenced views on independent artisans in the aftermath of the industrial revolution. Next, we discuss the positive and negative implications of the post-industrial view on AI generated art. Finally, we consider current issues raised by these models, and conclude by reflecting on the analogous issues seen in the industrial revolution.

\section{A Historical Shift}

\subsection{From Individual to Factory Worker}
The switch from an individualist working environment to a factory-centered one would permanently change the way society viewed independent artisans, bringing in a new age of commercialization that was predicated on the swift production of machine-made products over man-made ones. This difference on workforce style in is how the Industrial Revolution led to a change in “people’s relationship to craftsmanship, time, community and their own role in society as a whole”.

Such a change in working style was caused that it is hard to imagine a time where machines were not at the forefront of civilization. As the industry transformed resemble today's working world, small-scale artisans were pushed out of people's minds in favor of production that was faster and cheaper. The individual craftsman all but disappeared from the forefront of the business world. By the 1850s, most independent shoemakers had been replaced by shoe factories, independent weavers had gone out of business, and women with hand looms were quickly outstripped by the factories and machines bringing more people cheaper goods of higher average quality \cite{smail}.

However, a plethora of small-scale and skill-intensive sectors, like those in the metal trades and textile industries, managed to develop alongside the rise of factories \cite{berg}. Parts of the world also still value individually-made fine arts objects, especially cultures in the east like China and India. Berg points to the idea of luxury fashion in France or small-scale building restoration which is popular in Europe \cite{berg}. Though factorization still prevails, it is also important to note that the since the early 2000s demand for ``niche'' artisans has actually shown an upward trend. \citet{Alliance} reports that the global market for artisan-made products has increased by more than $8\%$ per year since 2002, and is worth more than $\$32$ billion. One of the reasons for such a trend has been an increased willingness to pay a premium for distinctive vis-à-vis mass-produced, goods.

\subsection{Machines Have Politics} 
\citet{winner} provides an illuminating example of transforming industry in “Do Artifacts Have Politics?” when he looks at the industrial mechanization of Chicago in the 1880s where the switch to factory production hurt skilled workers. When pneumatic molding machines were implemented by Cyrus McCormick’s reaper manufacturing plants, it drove individuals out of business \cite{winner}. In many other industries, this happened because the factory process was cheaper, but that was not the case here. In this instance, McCormick and the National Union of Iron Molders were at war, so even though the iron molding machines were not cheaper, they were used to push the previous workers away from unionization and out of business. Therefore, while the addition of industrial manufacturing hurt many skilled laborers naturally through cheaper replacements, it also hurt them unnaturally through the furthering of political agendas \cite{winner}. 

Unarguably, Winner concludes that the molding machine thus has politics in that its technical arrangements have become a form of order. Instead of subscribing to the use of pneumatic molding machines to speed up or cheapen production, these machines expressed human motives in their use towards achieving authority over others. It is undeniable that even if artifacts and machines do not have politics, they do indeed have power. 

In his famous essay ``The Work of Art in the Age of Mechanical Reproduction,'' \citet{benjamin1935work} discusses the change in perception of art in the age of mass production. Before mechanical reproduction, art was unique and valued for its ``aura'' -- which was derived from its authenticity and its physical and cultural context. However, with the ability to mechanically reproduce works of art, the traditional bases of cultural authority and hierarchy were challenged. As art moved to being based on politics instead of tied to rituals, it began to serve as a tool for political activism and resistance, ultimately bringing about social and political change \cite{benjamin1935work}.

Historically, it seems fair to say that the introduction of new technologies in the industrial revolution had many impacts on the individual craftsmen. Some of these impacts might be considered a natural course of action, but it is imperative as we move forward to acknowledge that other forms of use can be exacted through the introduction of new technology--use beyond that of merely what a machine physically produces.

\section{AI Art Generation}
We focus on the recent introduction of AI-generated art software to the current art world. AI-based art generators such as Dall-E \cite{dalle1, dalle2} and Midjourney are relatively new pieces of technology that can now create fully rendered images based solely on a user’s prompt, often producing impressive and intricate results.
If the industrial revolution changed the way society viewed artisans and craftsmen, how might the AI art revolution do the same? We now examine AI image generators as a computing technology that has the potential to cause this analogous shift in the art industry.

\subsection{The Perception of AI Art}
The most general fear associated with AI-generated art is that it could drastically reduce the amount of jobs available to working commercial artists today in areas such as illustration, animation, and graphic design. However, some others are concerned with the idea that the more ``traditional" notion of art may also be modified by AI-generated art. 

In ``The Culture Industry: Enlightenment as Mass Deception," \citet{adorhork} introduce the term ``culture industry," and compare technological advancement of mass media and creation to factory production of goods. They refer to the ``assembly-line character of the culture industry, the synthetic, planned method of turning out its products" as creating a passive society that is being manipulated into being satisfied by the products of capitalism, rather than by way of true psychological needs such as freedom, creativity and happiness \cite{adorhork}. 

In 2018, the art-collective Obvious, produced an art piece ``Edmond de Belamy," via a generative adversarial network (GAN) software package. The artwork was printed onto a canvas and sold at auction for $\$432,500$, over 43 times its pre-auction estimated value \cite{art}. Adorno and Horkheimer might argue that artwork created by an AI is created to be an ideal of art, and that AI artwork is merely feeding into mass-media culture industry that threatens ``high arts."

Oppositionally, the idea that AI-created art could be worthwhile in itself as an art piece, as seen in its high price valuation, points heavily towards James Moor's prediction that technology is shifting the questions we ask from ``How well does a computer do such and such an activity?” to ``What is the nature and value of such and such an activity?” \cite{moor}. In our case, the question shifts from ``How well can computers make art?" to ``What is art?".

\subsection{What is Art?}
The current literature on human attitudes towards AI-generated art presents some evidence as to what people might feel of this shift in viewpoint. In two recent studies \cite{Hong22,Kneer22} on attitudes towards artwork produced by humans versus by artificial intelligence, researchers found that while most test subjects felt that AI-generated art could be considered ``art," they were much less inclined to consider the art to be produced by an ``artist". This is a significant distinction because it implies that while artwork contains artistic value just by existing, artwork is not made just by putting pen to paper.

Thus we might push the shift Moor describes even further, from the question of ``What is art?" to ``What is an artist?". ``When judging whether an object falls under the category of `artwork,' the intent of the creator is seen as more important than even the appearance of the object in question" - \citet{Kneer22} seem to think, this is the reason that their participants were not unwilling to consider art created by a ``robot" to be ``art," but were significantly more at odds with calling a ``robot" an ``artist." 

In shifting the question from the object to the characteristic identity, we come to a yet unsolved question about AI art: If artwork generated by AI can be called ``art", but the model is arguably not an ``artist," then who is the author of such a piece? Mikalonyte and Kneer posit that this question has not yet been solved, suggesting that the lack of answer is reflected in the fact that autonomously generated AI artwork has yet to be copyrighted, with proposals to give copyright to the human designers of the artificial intelligence, as well as to redefine ``authorship" so as to include robots in the definition \cite{Kneer22}. 

\subsection{Human Art from Ends to Means}
Besides raising questions in moral philosophy, such an attribution to the functional aspects of art vis-à-vis the aesthetics could mean a lot of hope for traditional artists, especially those who were dependent on the techniques rather than the aesthetics of the final product. Artists who would only be differentiated by techniques might resort to promoting unusual techniques which are beyond the current scope of AI art models, or at least AI art models at present (e.g painting on paper towels or woodblocks, using the back of the paintbrush, using fingernails to paint). It wouldn't be surprising if artists would resort to differentiating factors relying on ``means'' rather than ``ends''. Artists would especially want their work to be intrinsically different - earlier that could be achieved via both unique propositions of outcomes and of methodology, but now it would largely be the latter. It won't be a surprise to witness more conservative forms being reinforced~\cite{browne2022or}. Analogous possible trends of increased interest in niche artworks as against mass-produced ones would actually further promote such resorting to traditional artistry and differentiating means.

\subsection{Fast-Paced Computational Creativity}

Pelz’s viewpoint that people became appendages to machines during the industrial revolution clearly has art analogues today. Over the years, a large section of the art industry has already resorted to computational methods~\cite{jenjac, feldman2017co} after witnessing the benefits of generative art. In generative art, new concepts, forms, shapes, colors, or patterns are created algorithmically. Artists or programmers first establish some criteria, post-which a computer creates new art forms adhering to those criteria. Such generative art is considered more aesthetically pleasing than functional \cite{r:19}. Hence wherever the functional aspects of art would be irrelevant, like in branding and advertising or the larger design sector, this transition towards computational art can accelerate people's dependence on art designs which can be quickly iterated. 

\subsection{Reduced Dependency on Traditional Artists} 
Suddenly, a piece of art that may have taken days to produce by a professional can be done with a handful of suggested words by almost anyone. It must be noted that if the emergence of this technology follows a similar path to that of mechanization following the industrial revolution, it could devalue commercial artists significantly and create massive job loss in an industry that was previously known to require a deeply human touch. Everything from storyboarding, concept art, and movie creation to advertising work and social media would be significantly different, causing a massive problem for the individual artists who may have trained for years to perfect their skill sets. 

\subsection{Benefits to Business}
Art has previously been far from a process one could automate, but these AI art generators might be the cause of a grand shift from skilled to unskilled labor in the free-lance art world. In the past few months, several online publications have tested using tools like Dall-E or MidJourney to provide art to accompany their written content \cite{Warzel22B}.

Access to AI art generation tools could be seen as an improvement to the current system for many business owners where there is a strong demand for commercial art. Being able to generate content for websites, branding, marketing and sales in a virtually cost-free manner could help small-business owners to reach bigger audiences. Online publications have largely stepped away from free-lance artists anyways, with many publications hosting content that was created elsewhere (i.e. embedded tweets or stock photography).

Stock photography businesses sell royalty-free pictures for personal or commercial use, usually paying $15-40\%$ to the creator of the photo for each license \cite{photo2}. Recently, AI have been breaking into this market, with the distinct art styles of Dall-E or Midjourney popping up on stock photo websites \cite{photo}. Shutterstock, one of the largest stock photo retailers, announced that along-with OpenAI, they would be banning artwork from other AI generators from being uploaded to their site, and it would also create a “Contributor Fund” to help pay the artists whose work was used to train the AI software \cite{photo2}.

\subsection{Art Democratization} 
Platforms like YouTube, Instagram and TikTok, which have become hosts of content creation, have been able to attract millions of content creators who make money using their services by providing them with fame and monetary rewards. This has largely happened with the reduced technical and social barriers that these arguably democratic platforms have provided. AI art models could also tread the same path. Democratization of art would mean almost anyone can produce artistic creations, including a person without limbs or someone with a neurological disorder that affects their ability to draw or paint. AI-art models might be exclusively seen as potential attackers on the most talented segments of the artistic society, but they will doubtlessly open up a level playing field for those who considered art out of reach. Besides, such democratization would also be reflected in crowd-sourced efforts~\cite{crowdsourcingandcrowdwork,kittur2013future,kittur2019scaling,dhole2021nl,srivastava2022beyond} which would seek contributions to aid in developing large creative models fairly.

\subsection{Increased Amount of Plagiarism}
The Industrial Revolution was replete with examples of industrial espionage \cite{harris1985rolt, plagiarism} where large businesses often stole the work and ideas of people who had historically performed it. Governments regularly encouraged individuals to steal ideas, especially from abroad, since it hardly required applicants to be inventors, especially if the invention was abroad. The current AI art models already have been exposed to the artworks of many artists without giving them proper attribution or even seeking their permission. Millions of generations of artwork have already been utilizing these styles. Besides artists’ work being plagiarized, it is unclear how credit would get divided amongst the artists, model trainers and users writing prompts. While there have attempts to legally delineate the complete generation pipeline~\cite{fjeld2017legal, kim2020ai}, the black box nature would make it hard for fair credit attribution.  

\subsection{Increased Carbon Emissions}
The dramatic increase in coal and gas usage, which skyrocketed pollution levels across major cities and industrial zones, was an unfavorable side consequence of the Industrial Revolution. Today, with large groups of people expected to move towards careers of computational art, it would be inevitable that training these AI-art models would also be performed more frequently, via researchers as well as artists raising concerns of carbon emissions.~\citet{strubell2020energy}'s lifecycle assessment of training popular large language models revealed that a typical training process took nearly five times the lifetime carbon emissions of the average American car. Besides, the largeness of these models also necessitates GPU usage during inference time.

\subsection{Furthering Political Agendas}
\citet{winner} has been helpful for giving convincing arguments that artifacts and machines in general can have biases. The 9ft clearance levels of Long Island bridges were a design decision made by urban planner Robert Moses in order to restrict buses filled with low-income people and racial minorities from accessing parkways. As a result of the biases implicit in these designs, people were given limited access to parts of their own city \cite{winner}.

\citet{benjamin1935work} had argued that as art's authenticity diminishes due to the ease of mechanization, it begins to be based on politics rather than on rituals. Just like the Long Island bridges, political agendas intertwined with design have had crucial consequences. Therefore, it wouldn't be a surprise if biases mirrored in the AI art generation of today~\cite{bansal2022well} were exploited to further political agendas. ~\citet{Hassine} revealed that AI generated art skews mostly white, both in depiction and in representation. Unless age, sex or race is specified, prompts to the system have built in biases towards young white men \cite{biases}. This unconscious bias could have a manifestation in the real world, just as Robert Moses' designs manifested for racial minorities in New York City.

\section{Questions of Concern}
Finally, we consider the issues to pubic welfare and society that AI-art generation introduces through its effects on the artists of today. We also consider the scope of involvement that computer scientists and AI have had in creating or contributing to these issues.

\subsection{Is the threatened change in the status of artists characterized by the primary and essential involvement of AI models?}

Industrial “sweatshops” mass-producing art for commercial consumption have been a constant long before the computer became an element in the equation. From the comic strips of the 1880s - 1960s to the comic books of the 1930s - present day, to the mass produced landscape art created for furniture stores in Asian factories since the mid-1950s, art has been commoditized long before the computer \cite{comic}. In these assembly lines, one person would sketch the outlines of image, another would pencil in details of the people, another would ink those images, yet another would draw in the backgrounds, and a final hand would color the image. Produced by an assembly line usually called a “studio,” the art would be signed either by an arbitrarily selected worker or even by a completely fictitious artist \cite{comic}. More to the point — this was art created by “factory workers” who acted the same role in production as today’s graphics programs do \cite{movie}. Therefore, one could also argue that AI models might not be essential to the problem. However, what distinctively stands out with the usage of AI art models, is the rapid pace of artwork creation and proliferation, unlike what was witnessed before the arrival of computers or the internet.
 
\subsection{Does the threatened change in the status of artists occur because of exploiting some unique property of AI models?}

The primary difference created by technology is mass access. To staff a “studio” with bit-work artists requires a substantial investment in infrastructure, equipment, and labor costs. Such programs as are available today are much more economical and widely available to individuals — from hobbyists to serious artists to mercenary corporations — than at any time in our history. So it is legitimate to argue that AI models have uniquely driven the scale and creativity of the problem far more broadly than earlier technology could have.
 
\subsection{Could this issue have even arisen without the involvement of AI models?}

The answer depends heavily on the question of whether AI-art generation programs are considered as tools or entities. Motion-picture technology created entirely new art forms. One could create art with motion across space and time in a way that entirely changed how our culture thought about visual art. But the cameras themselves have never — for all their technological sophistication — been more than tools. Cameras do not set out with purpose to make movies, and AI art generation programs do not set out independently to create images. Both must be employed by users.
 
\section{Conclusion}
 
Perhaps what AI-art generation software is doing is forcing our society to confront a much larger issue about artists. Instead of threatening the status of artists, advances in computer technology require us to confront the idea that the view of artists has already changed, and has been changing. 

While on the surface AI-generated art seems unique, many of the issues that it is raising concerning society's views of art and artists are merely more complex callbacks to the mechanization of artisan's projects in the industrial revolution. Specifically, the art generated by AI-generation algorithms creates problems that mirror those of industrialization. In the same way that the artisan was pushed out, so too is the artist today. In the same way intellectual property was compromised during the 1800s, legal loopholes may ensure many artists are not duly credited without proper AI regulation. 

Our study of the industrial revolution analogues serves as a warning to look backwards at the past treatment of creators and consumers of industry in the wake of newly introduced technologies. We pose that this may be an important step to take before experiencing the consequences of the technical revolutions that unfold before us today.

But these analogies should not be taken as a discouragement against developing large models or to undermine the efforts of the field of AI in general. We should actively strive to improve technical parameters of these models, by accounting for the possibility of potential damage early on, as these models have and already display tremendous potential for business as well as for democratization. 

\section{Limitations}
How AI art generation tools will affect artists is an extremely subjective and multifaceted subject, and forecasting it precisely will not be easy. We have provided comparisons based on events that occurred post the industrial revolution. However, we think that empirical evidence would be helpful in better understanding many of the issues raised. Our objective was to present as thorough and comprehensive an analysis as possible by considering the technical, political and industrial implications of art. Our section on ``What is Art?’’ is quite limited due to the rich history concerning this question from a philosophical point of view, but we still feel it was important to include. Nonetheless, we believe that our work will serve as a crucial gateway for both the engineering and humanities disciplines to facilitate dialogue and advance debate about the impact of AI art tools.

\section{Acknowledgments}
We thank Dr. Kristin Williams and Dr. Steve Newton for their crucial thoughts and feedback in numerous drafts. We also thank the anonymous reviewers and meta reviewer for their invaluable suggestions.

\bibliographystyle{aaai22} 
\bibliography{aaai22}

\end{document}